# A LSTM and Cost-Sensitive Learning-Based Real-Time Warning for Civil Aviation Over-limit

Yiming Bian
China Jiliang University
Hangzhou, China
alexbian@cjlu.edu.cn

*Abstract*—The issue of over-limit during passenger aircraft flights has drawn increasing attention in civil aviation due to its potential safety risks. To address this issue, real-time automated warning systems are essential. In this study, a real-time warning model for civil aviation over-limit is proposed based on QAR data monitoring. Firstly, highly correlated attributes to over-limit are extracted from a vast QAR dataset using the Spearman rank correlation coefficient. Because flight over-limit poses a binary classification problem with unbalanced samples, this paper incorporates cost-sensitive learning in the LSTM model. Finally, the time step length, number of LSTM cells, and learning rate in the LSTM model are optimized using a grid search approach. The model is trained on a real dataset, and its performance is evaluated on a validation set. The experimental results show that the proposed model achieves an F1 score of 0.991 and an accuracy of 0.978, indicating its effectiveness in real-time warning of civil aviation over-limit.

*Keywords—QAR data, LSTM, cost-sensitive learning, grid search method, Spearman rank correlation coefficient*

## I. Introduction

QAR (Quick Access Record) data is a key component of aviation safety data, which records the aircraft's flight parameters including G-value, pitch angle, etc. These data are obtained by a large number of sensors. These data are acquired by a large number of sensors. By setting certain threshold values for the parameters recorded by QAR, when the parameters in flight exceed the threshold values, QAR over-limit occur and are recorded accordingly [1]. According to the iceberg theory and Hain's law, although these exceedances do not necessarily lead to serious consequences at the moment, they are likely to lead to serious accidents and safety hazards after years of accumulation. As a support, the International Civil Aviation Organization (ICAO) has required its Contracting States to conduct safety hazard studies for minor consequences. It can be seen that the risk of serious aviation accidents can be effectively reduced by monitoring various QAR data during flight and providing timely warning of potential over-limit events. Therefore, while strengthening the monitoring of QAR data, further research on the change pattern of QAR data in time, mastering the evolution pattern of QAR data over-limit and other characteristics, and establishing an automated early warning model of flight over-limit in civil aviation will be of great significance in moving forward the safety gate of civil aviation flight safety management.

Many scholars have conducted a lot of research on QAR over-limit events, mainly focusing on the following two aspects. One is to use QAR data to predict future flight parameters, for example, Chao Tong [2] et al. proposed to predict the vertical acceleration (VRTG) at the moment of aircraft landing using LSTM recurrent neural network; Kang, Z. [3] et al. proposed to predict aircraft landing speed sequence based on QAR data. Second, QAR data are used to diagnose and analyze the causes of accidents such as over-limit, for example, W. Xiangzhang [3] et al. combined hard landing cases and QAR data to build a SPA-Markov model to analyze the risk of hard landing for a fleet of aircraft in each month; F. Tian [1] et al. based on QAR data, for the already controlled flight into terrain events based on QAR data. All of the above studies have achieved some results, but there are fewer studies that use QAR data as a basis for real-time warning of flight over-limit.

With the development of artificial intelligence and machine learning, these techniques are increasingly applied in various engineering fields. RNN is one of the deep learning models, which shows extremely strong adaptability in time series data analysis. Long short-term memory network (LSTM) is a variant of RNN, which effectively solves the problems of gradient disappearance and gradient explosion of RNN and enables recurrent neural networks to effectively utilize long-range temporal information. LSTM has been successfully applied to temporal data analysis in several fields, including Natural language processing systems [5] , Image analysis [6], time series prediction [7], and failure time series prediction [8], etc. It is easy to see that the LSTM model can also be applied to the civil aviation over-limit warning problem, which is also time series prediction. It can be used to determine whether an over-limit accident will occur in the future through real-time detection and storage of various flight parameters of the aircraft to achieve automated early warning.

In this paper, an automated warning model based on long short-term memory recurrent neural network (LSTM) for civil aviation over-limit is established to realize automated warning by detecting each flight parameter of the aircraft in real time and determining whether a safety accident is currently occurring. The work in this paper is mainly the following steps: firstly, the data set is pre-processed by using data coding and adjusting the sampling frequency. Secondly, the QAR parameters with high correlation with the over-limit are extracted by using Spearman correlation analysis. Finally, an LSTM model is built to realize the prediction of the overlimit, and the grid search algorithm is used to prefer the overparameters of the model. In this paper, experiments are conducted on an Internet public dataset to verify the early warning capability of the model. The experiments show that the model proposed in this paper performs well in terms of accuracy and F1 score for an imbalanced binary classification problem like flight over-limit.

## II. LSTM MODEL BUILDING

### A. Loss Function

This paper deals with a binary classification problem, so the following Binary Cross Entropy Loss can be used as a loss function and as a performance evaluation metric for the model:

$$BCE = -\frac{1}{n}\sum_{i=1}^{n}\left[y_i \cdot \log p(y_i) + (1-y_i) \cdot \log\left(1-p(y_i)\right)\right] \quad (1)$$

where $y_i$ is the binary attribute value 0 or 1, and $p(y_i)$ is the probability that the outcome predicted by the model belongs to the attribute value $y_i$.

Since there are very few moments of over-limit in the flight data of a flight compared to the normal flight moments, it can lead to a huge difference in the number of samples between the positive class (normal) and the negative class (over-limit). For this kind of class imbalanced binary classification problem, cost-sensitive learning, undersampling and oversampling can be used to reduce the impact of uneven sample distribution on the model performance. In mainstream classification models such as ANN, cost-sensitive has been widely used. In recent years, with the continuous in-depth research on deep learning, the idea of cost-sensitive learning has also been incorporated into it. For example, Chung et al [9] introduced the cost factor into the loss function of CNN to improve the recognition accuracy of traditional CNN for few classes of samples. In this paper, based on the binary cross-entropy loss function, the cost matrix is combined.

$$COST = \begin{bmatrix} 0 & cost_{01} \\ cost_{10} & 0 \end{bmatrix} \quad (2)$$

where $cost_{ij}$ is the cost of predicting the sample of class $i$ to the sample of class $j$. And when $i = j$, $cost_{ij} = 0$. In this paper, defining class 0 as positive cases and class 1 as negative cases, then $cost_{01}$ is the percentage of positive cases in the training set and $cost_{10}$ is the percentage of negative cases in the training set. Thus, the cost-sensitive binary cross-entropy is obtained as follows:

$$loss = -\frac{1}{n}\sum_{i=1}^{n}\Big(y_i \cdot \log p(y_i) \times cost_{01} + (1-y_i) \\ \cdot \log\left(1-p(y_i)\right) \times cost_{10}\Big) \quad (3)$$

Equation (3) is used as the loss function in this paper.

### B. LSTM structure

Compared with the basic RNN model, the LSTM model overcomes the problem of gradient explosion and gradient disappearance by introducing three gating units: an input gate, a forgetting gate and an output gate. These three gating units are used to control the input of new information, control whether to discard the previous information, and control the output information, respectively. In addition, the LSTM has a memory unit and a candidate memory unit that can be turned into a new memory unit [2].

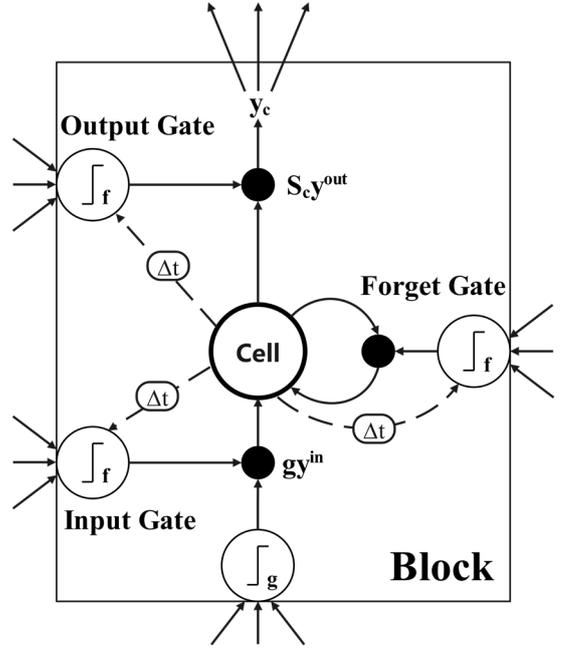

Fig. 1. Schematic diagram of LSTM memory block

The structure of the LSTM neural network is briefly represented in Fig. 1. This structure gives the LSTM model a better ability to handle long sequence data.

### C. LSTM-based over-limit warning model

The LSTM early warning model established in this paper consists of one input layer, two hidden layers and one output layer, as shown in Fig. 2. Among them, the two hidden layers are fully connected layers. Specifically, the input of each neuron is the output of all neurons in the previous layer, i.e., each element of the input vector is multiplied with the corresponding row in the weight matrix and the results are summed to obtain a weighted sum. The QAR data of the input model is denoted as X, which is a 3-dimensional array of $i \times j \times k$. I is the number of samples used by the model for one training. j is the sequence length of the input data, and k is the number of features of the input data. The output of the model is y, which is an i-dimensional vector and $y_i \in \{0,1\}$.

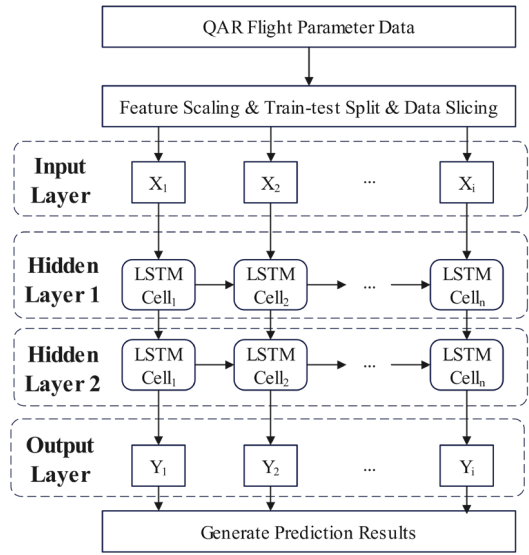

Fig. 2. A framework of LSTM-based flight over-limit warning model

The training process of LSTM model is similar to that of neural network model. The updates of the states at time t can be written as [10]:

$$net_{in_j}(t) = \sum_m w_{in_j m} y^m(t-1) + \sum_{v=1}^{S_j} w_{in_j c_j^v} s_{c_j^v}(t-1) \quad (4)$$

$$net_{\varphi_j}(t) = \sum_m w_{\varphi_j m} y^m(t-1) + \sum_{v=1}^{S_j} w_{\varphi_j c_j^v} s_{c_j^v}(t-1) \quad (5)$$

$$s_{c_j^v}(0) = 0 \quad (6)$$

$$s_{c_j^v}(t) = y^{\varphi_j}(t) s_{c_j^v}(t-1) + y^{in_j}(t) g\left(net_{c_j^v}(t)\right) \quad (7)$$

$$net_{out_j}(t) = \sum_m w_{out_j m} y^m(t-1) + \sum_{v=1}^{S_j} w_{out_j c_j^v} s_{c_j^v}(t-1) \quad (8)$$

where $y^m(t) = f_m(net_m(t))$, $m$ can be forget gate "$\varphi$", input gate "$in$" and output gate "$out$", respectively. "$\Delta t$" refers to the peephole connection [10].

$$y^{in_j}(t) = f_{in_j}\left(net_{in_j}(t)\right) \quad (9)$$

$$y^{\varphi_j}(t) = f_{\varphi_j}\left(net_{\varphi_j}(t)\right) \quad (10)$$

$$y^{out_j}(t) = f_{out_j}\left(net_{out_j}(t)\right) \quad (11)$$

$$y^{c_j^v}(t) = y^{out_j}(t) s_{c_j^v}(t) \quad (12)$$

The suffix $c$ refers to an element in the set of cells CEC. "$s_c$" refer to the state value of cell c – i.e. its value after the input and forget gates have been applied, $f, g, h$ are the squashing functions of the gates, the cell input and output, respectively, "$w_{ij}$" is the weight $from$ unite $j$ to unite $i$ [10].

The model enables early warning of whether an over-limit occurs at time t by inputting QAR data from (t-j) to (t-1) moments.

*D. Training of over-limit warning model*

In the input layer, let the input QAR data be **X** and the sequence corresponding to whether an over-limit occurs at the current moment be **Y**. MinMax normalization is applied to normalize **X** to obtain **X'**:

$$X'_{ij} = \frac{X_{ij} - \min(\mathbf{X_j})}{\max(\mathbf{X_j}) - \min(\mathbf{X_j})} \quad (13)$$

Meanwhile, 80% of the data in X' and Y are randomly divided into training set $\mathbf{X_{tr}} = \{X'_1, X'_2, ..., X'_m\}$ and $\mathbf{Y_{tr}} = \{Y_1, Y_2, ..., Y_m\}$, and the rest are test sets $\mathbf{X_{te}} = \{X'_{m+1}, X'_{m+2}, ..., X'_n\}$ and $\mathbf{Y_{te}} = \{Y_{m+1}, Y_{m+2}, ..., Y_n\}$.

The time step of the LSTM model is set as T, the number of LSTM units in each hidden layer is n, and the learning rate is $\alpha$. Equation (3) is used as the loss function, and the LSTM model is optimized using the Adaptive Moment Estimation algorithm.

*E. LSTM model hyperparameter optimization based on grid search*

LSTM over-limit warning model established above, many hyperparameters in the model are involved, including time steps (T), LSTM units (n), and learning rate ($\alpha$). In practical use, appropriate hyperparameters need to be selected according to the specific task and data set to obtain the best model performance and generalization ability. In order to improve the model performance, this paper uses the grid search method to optimize these three hyperparameters. The grid search algorithm is a simple and effective method to find suitable hyperparameters. It calculates the average accuracy of the model by traversing all the pre-defined hyperparameter fetches and performing k-fold cross-validation in each set of different hyperparameter combinations, i.e., training the model on the training set and testing the model performance on the validation set at the same time. The objective of grid search hyperparameters is to minimize the average error $\bar{\varepsilon}$ of hyperparameter combinations on the test set, and the objective function is:

$$\min \bar{\varepsilon} \quad (14)$$

The constraints are:

$$s.t. \begin{cases} T \in \{T_1, T_2, ..., T_i\} \\ n \in \{n_1, n_2, ..., n_j\} \\ \alpha \in \{\alpha_1, \alpha_2, ..., \alpha_k\} \end{cases} \quad (15)$$

The hyperparameters $T, n, \alpha$ to be searched constitute a three-dimensional search space, and the search process is shown in Algorithm 1. First, each time step value is searched, i.e., the data is sliced into sizes corresponding to the length of time. Then a two-dimensional search space consisting of $n, \alpha$ is constructed using the GridSearchCV function in the sklearn library, and the LSTM model constructed in the previous section is trained. A k-fold cross-validation method is used to calculate the average accuracy of the model under different combinations of hyperparameters. After traversing all combinations, the preferred hyperparameter combination is obtained. In this paper, time steps $T \in \{3, 5, 7, 10, 30, 50\}$, LSTM units $n \in \{10, 20, 30, 40, 50, 60, 70\}$, and learning rate $\alpha \in \{0.001, 0.003, 0.005, 0.007, 0.01, 0.03\}$ are set.

---

**Algorithm 1: Hyperparameter tuning for LSTM model**

**In put:** D, epochs, k
**Out put:** The hyperparameter combination with the highest accuracy on the test set
**Predefine:** value of epochs, k
**Predefine:** value ranges of $T, n, \alpha$
**for** each $T$ in $T_1, T_2, ... T_i$
  **for** each $n$ in $n_1, n_2, ... n_j$
    **for** each $\alpha$ in $\alpha_1, \alpha_2, ... \alpha_k$
      execute LSTM model
      append results with $T, n, \alpha, \bar{\varepsilon}$
    **end for**
  **end for**
**end for**
**return** results ranked by $\bar{\varepsilon}$

---

III. EXPERIMENTAL EVALUATION

In this section, the LSTM over-limit warning model proposed in Section 2 and its hyperparameter preference algorithm are combined with real civil aviation QAR data for the validation of the experiment.

*A. Data set pre-processing*

The data set used in this paper is the QAR data of eight flights throughout the flight [11], which contains about 205,000 samples and their 60 attributes, recording the flight parameters of passenger aircraft during the flight. In this paper, the data types and data contents in the dataset are first analyzed, i.e., the text labels are converted into numerical labels, then the sampling frequency of each data is uniformly adjusted to 1 second, and finally the data is labeled and the key data is filtered out. Finally, the data are labeled and the key data are selected for subsequent training and evaluation of the model.

*1) Convert attributes from text to values*

In the data set, some attributes values are represented by text labels, such as DOWN/NaN, FALSE/TRUE, DISENGD/ENGAGED, etc. In order to facilitate subsequent data analysis and processing, this paper converts these text labels into numerical labels. Specifically, according to the actual meaning of each label, it is converted into the corresponding numerical value. The conversion results are shown in Table 1 below:

TABLE I. THE RESULT OF CONVERTING TEXT LABELS TO NUMERIC LABELS

| Attribute | Original attribute value | New attribute value |
|---|---|---|
| GEAR SELECT | NaN | 0 |
| DOWN | DOWN | 1 |
| WOW INDICATE | FALSE | 0 |
| INAIR | TRUE | 1 |
| A/T ENGAGED | DISENGD | 0 |
| | ENGAGED | 1 |
| ANY A/P | ON | 0 |
| ENGAGED | OFF | 1 |

*2) Sampling frequency adjustment*

Some attributes recorded multiple sets of data in 1 second. To facilitate data processing and modeling, the mean (or plurality) of each attribute within 1 second is calculated in this paper, and the sampling frequency is adjusted to 1 time/second. This method can effectively reduce the amount of data while retaining important information of the data, making data analysis and modeling more accurate and reliable. The specific process is as follows:

*a) For attributes with continuous eigenvalues*

For an attribute with the same name that originally had multiple consecutive sets of attribute values in 1 second, such as the COG NORM ACCEL attribute, the sampling frequency is now adjusted to 1 time/s, and its new attribute in the ith second is obtained after taking its average value in every 1s second.

*b) For attributes whose feature value takes the value 0 or 1*

For attributes with the same name that had j groups of values equal to 0 or 1 in the i-th second, such as the WOW INDICATE INAIR attribute, the sampling frequency is now adjusted to once per second. This means that the new attribute value for the i-th second is obtained by taking the mode of the values within each one-second interval.

The new data set is denoted as X. This process achieves a reduction in the number of attributes in the data set, thus reducing the dimensionality of the data to a certain extent.

*3) Labeling of over-limit data*

Since the QAR is not marked within the dataset whether the limit is exceeded or not, the Laida criterion combined with the 10,000 flight rate triggered by historical QAR exceedance events is used to determine the warning criteria for each indicator by combining the G values in the QAR data [3]. The calculation formula is:

$$\sigma_G = \sqrt{\frac{\sum_{i=1}^{n}(G_i - u_G)^2}{n}} \quad (16)$$

where $G_i$ is the G-value of the ith sample, $u_G$ is the average G-value of all samples, and n is the number of samples.

$G_i$ all satisfy $G_i \geq 0$, that is, $G_i$ indicates the magnitude of acceleration. Therefore, for $G_i \geq u_G + 3\sigma_G$, it is labeled as over-limit data ($y_i = 1$), and for $G_i < u_G + 3\sigma_G$, it is labeled as normal ($y_i = 0$). From this, the sequence $\mathbf{Y} = (y_1, y_2, \ldots y_n)$ with the meaning of whether flight over-limit occurs can be obtained, and $y_i = 0 \, or \, 1$.

*4) Key Feature Selection*

There are still 35 features in the dataset after the above steps of processing. In order to scientifically and effectively select the flight parameters with high correlation with the over-limit event, this paper adopts the method of calculating the Spearman rank correlation coefficient [12] among different features, which has the advantages of not requiring the data to satisfy a normal distribution, etc., and is especially suitable for the data set containing binary attribute values in this paper. The Spearman's rank correlation coefficients between two sets of eigenvalues $\mathbf{X_j} = (X_{1j}, X_{2j}, \ldots, X_{nj})$ and $\mathbf{Y} = (y_1, y_2, \ldots y_n)$ can be calculated by Equation (17):

$$r_s = \frac{\sum_{i=1}^{n}(p_i - \bar{p})(q_i - \bar{q})}{\sqrt{\sum_{i=1}^{n}(p_i - \bar{p})^2}\sqrt{\sum_{i=1}^{n}(q_i - \bar{q})^2}} \quad (17)$$

where $r_s$ is the rank correlation coefficient, and $p_i$ and $q_i$ are the rank of $X_{ij}$ and $y_i$, respectively. If equal values occur in the variables, the rank corresponding to that value is the average of the ranks corresponding to these values. If $0 < r_s \leq 1$, it indicates a positive correlation between the two sets of eigenvalues; if $-1 \leq r_s < 0$, it indicates a negative correlation; when $r_s = 0$, it means no correlation. According to the correlation analysis results, this paper selects a threshold value of $|r_s| > 0.05$, and selects nine flight parameters with high correlation with flight over-limit, as shown in Table 2.

TABLE II. RESULTS OF SPEARMAN RANK CORRELATION COEFFICIENT CALCULATION

| Attribute | Description | $r_s$ | P value |
|---|---|---|---|
| COG NORM ACCEL | Center of Gravity Normal Acceleration | 0.203 | 0.000 |
| CAP CLM 1 POSN | Captain's Column Position | 0.125 | 0.000 |
| GROUNDSPEED | Ground Speed | -0.115 | 0.000 |
| FMF GROSS WEIGHT | Weight of Aircraft | -0.093 | 0.000 |
| PITCH ATT RATE | Pitch Attitude Rate | 0.092 | 0.000 |
| COMPUTED AIR SPD | Calculated Airspeed | -0.078 | 0.000 |
| WIND SPD(ADIRU) | Wind Speed | -0.076 | 0.000 |
| PITCH ATT | Pitch Attitude | 0.073 | 0.000 |
| ANY A/P ENGAGED | Any Autopilot Engaged | -0.054 | 0.000 |

*B. Experimental platform and environment*

The configuration of the computer used for the experiments is as follows: CPU is intel CORE CPU i7-11800H, CPU frequency is 4.6GHz, memory is 16GB. GPU is RTX3060Laptop, video memory is 6GB. operating system is Windows 11 (64-bit); programming language is Python 3.8.0 (64-bit); The IDE is PyCharm Professional Edition 2023.1. The LSTM model was implemented by tensorflow-gpu 2.6.0 [13] and keras 2.6.0 framework during the programming process.

*C. Results and Analysis*

  *1) LSTM model training results*

Considering that the output of the model has only one boolean variable, i.e., whether the aircraft exceeds the limit at the next moment, the output layer of the LSTM model is set to one neuron in this paper. The remaining hyperparameters are initially set according to experience, time steps is set to 10, learning rate is set to 0.005, the number of LSTM units in the hidden layer is set to 30, and the size of epochs is set to 200.

As can be seen in Fig. 3, the LSTM model built in this paper has achieved better results at a training round of 40. At this time, the model has an accuracy of about 0.96 on both the training and test sets.

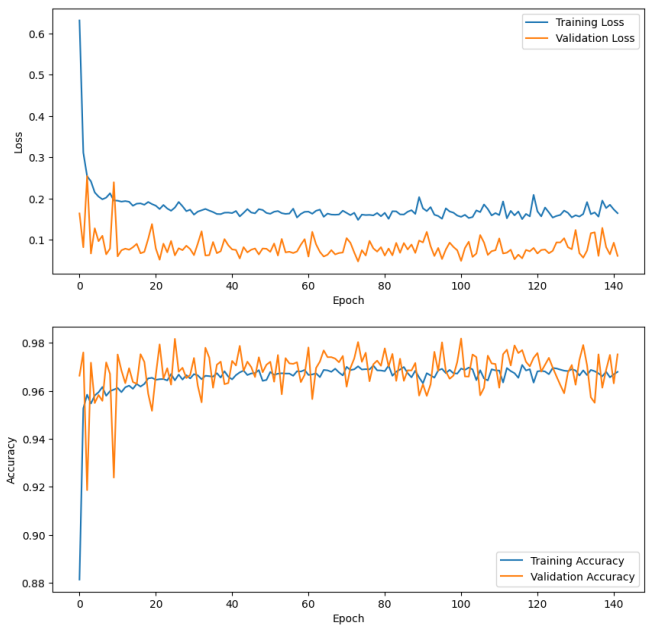

Fig. 3. LSTM model training results

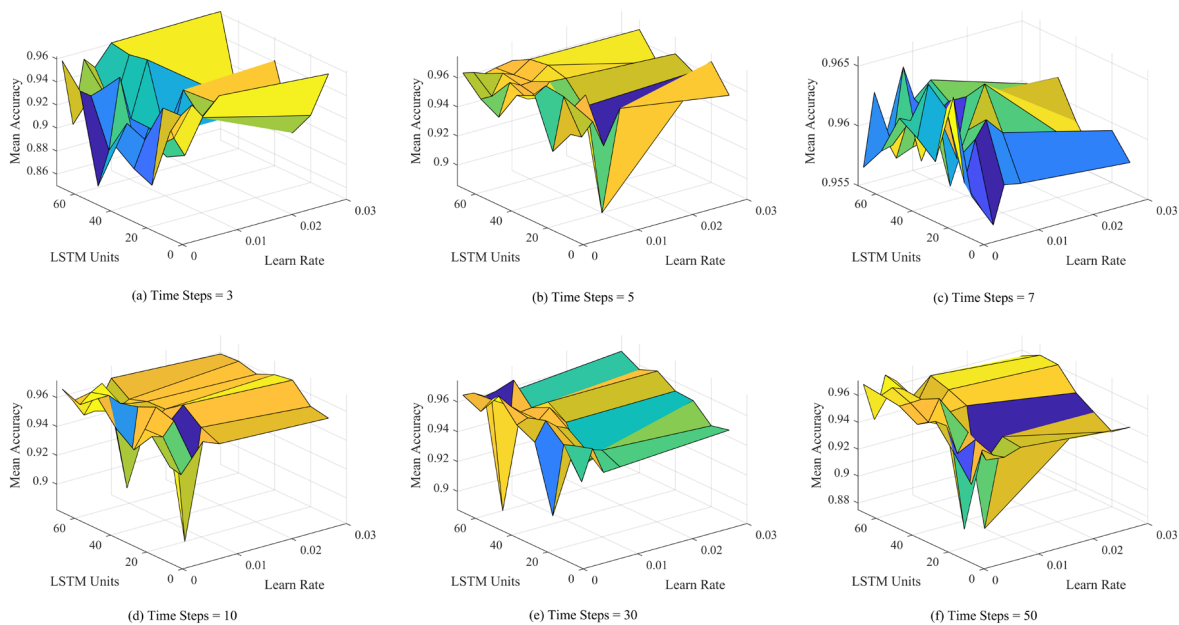

Fig. 4. Grid search results for LSTM model's hyperparameter

TABLE III. THE FIRST FIVE GROUPS OF OPTIMAL GRID SEARCH RESULTS FOR THE COMBINATION OF HYPERPARAMETERS OF THE LSTM MODEL

| Rank | Time Steps | LSTM Units | Learning Rate | Average accuracy | Average Fit Time Consuming (s) |
|---|---|---|---|---|---|
| 1 | 5 | 30 | 0.003 | 0.9746 | 77.7472 |
| 2 | 30 | 50 | 0.001 | 0.9741 | 111.3098 |
| 3 | 30 | 60 | 0.007 | 0.9733 | 111.5381 |
| 4 | 5 | 20 | 0.003 | 0.9730 | 75.3567 |
| 5 | 10 | 30 | 0.005 | 0.9727 | 84.6631 |

  *2) Results of hyperparameter optimization of LSTM models*

The grid search algorithm (Algorithm 1) mentioned earlier is used here to perform the optimal selection of hyperparameters in the LSTM model. First, the values of the non-critical hyperparameters are set, epochs are set to 5 and $k = 4$ in k-fold cross-validation. Then, the range of values of the three hyperparameters is set to find the optimal combination: time steps $T \in \{3,5,7,10,30,50,70,90\}$, LSTM units $n \in \{10,20,...,70\}$, and learning rate $\alpha \in \{0.001, 0.003, 0.005, 0.007, 0.01, 0.03\}$. Finally, the objective function is set to have the highest average accuracy in k-fold cross-validation.

In this paper, the results of grid search for LSTM model hyperparameters are shown in Fig. 4 by drawing 3D grid plots. Among them, the same graph shows the results for the same time steps in the hyperparameter combinations, the Z-axis represents the model accuracy, and the X,Y-axis represents the combinations of LSTM units and learning rate at different times. Table 3 shows the average accuracies in the top 5 optimal parameter combinations and their corresponding k-fold cross-validation for the dataset used in this paper.

The combination of hyperparameters with the highest average accuracy of the model under 4-fold cross-validation in grid search was selected: Time Steps = 5, LSTM Units = 30, Learning Rate = 0.003 to train the LSTM model, and the training results are shown in Fig. 5. By comparing with Fig. 3, it is obvious that the performance of the model has improved significantly after the hyperparameter preferences, and the model has an accuracy of more than 0.98 on both the training and test sets when epoch = 100.

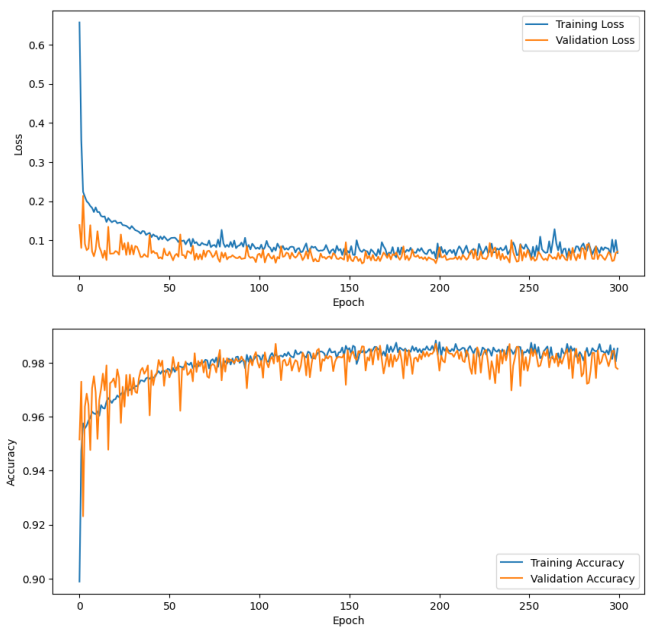

Fig. 5. Model training results after hyperparameter optimization of LSTM model using grid search

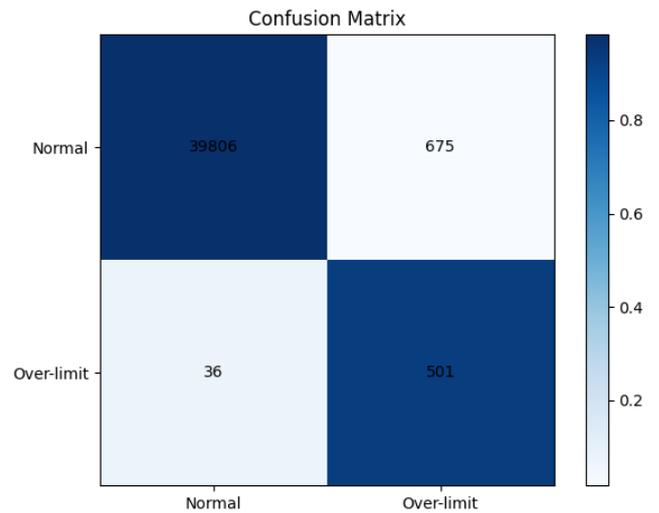

Fig. 6. Confusion matrix of the hyperparametric optimized LSTM model on the validation set

As Fig. 6 shows the confusion matrix of the model on the test set, combined with the previous definition of normal as a positive case and over-limit as a negative case, the Precision of the model can be calculated as 0.999, recall as 0.983, and F1 score as 0.991. And conclude that the model has good performance.

IV. CONCLUSION

Flight over-limit as a non-negligible problem in civil aviation transportation industry has caused great disturbance to ensure the safety of air transportation. In this paper, an automated warning model of civil aviation over-limit is proposed based on LSTM recurrent neural network and combined with cost-sensitive learning to predict whether the over-limit occurs in the next moment by analyzing the time series QAR data of flights in the past and current moments, and realize the real-time warning of over-limit. Meanwhile, this paper contains the training, evaluation and over-parameter optimization of the model. After inputting the pre-processed data into the model, the accuracy of the model is above 0.97 in the training set and validation set, and the F1 score is 0.991 in the validation set.

In the future, aviation-related fields can start by analyzing the QAR parameters related to flight over-limit, studying the key features of flight over-limit, and further optimizing the model. Meanwhile, the number of samples in the dataset used for training the model can be expanded to include QAR data of different aircraft types, etc., to analyze and improve the generalization of the model.